\documentclass[a4paper]{article}

\usepackage{INTERSPEECH2020}
\usepackage{amsmath}
\usepackage{amsbsy}
\usepackage{amssymb}
\usepackage{multirow}

\newcommand{\nc}{\newcommand}

\def\lb{\label}
\def\erf#1{(\ref{#1})}

\def\bx{\mbox{$\mathbf{x}$}}
\def\by{\mbox{$\mathbf{y}$}}

\def\bw{\mbox{$\mathbf{w}$}}

\def\bOmega{\mbox{$\mathbf{\Omega}$}}

\def\Y{\mbox{$\mathcal{Y}$}}

\long\def\gobbleup#1{}

\def\argmax{\operatornamewithlimits{argmax}}
\def\argmax{\mathop{\rm argmax}}

\nc{\bwq}{\mbox{$\bw_{\text{q}}$}}

\def\conjY0{\mbox{$\left[Y_n^m(\bOmega_0)\right]^*$}}
\def\Y0{\mbox{$Y_n^m(\bOmega_0)$}}

\def\sigs0{\mbox{$\sigma_{\text{sp},0}^2$}}

\title{Sequence-level self-learning with multiple hypotheses}
\name{Kenichi Kumatani, Dimitrios Dimitriadis, Yashesh Gaur, Robert Gmyr, \\Sefik Emre Eskimez, Jinyu Li, Michael Zeng}
\address{Microsoft, WA, USA}
\email{}

\begin{document}
\ninept

\maketitle
\begin{abstract}
In this work, we develop new self-learning techniques with an attention-based sequence-to-sequence (seq2seq) model for automatic speech recognition (ASR). For untranscribed speech data, the hypothesis from an ASR system must be used as a label. However, the imperfect ASR result makes unsupervised learning difficult to consistently improve recognition performance especially in the case that multiple powerful teacher models are unavailable. In contrast to conventional unsupervised learning approaches, we adopt the \emph{multi-task learning} (MTL) framework where the $n$-th best ASR hypothesis is used as the label of each task. The seq2seq network is updated through the MTL framework so as to find the common representation that can cover multiple hypotheses. By doing so, the effect of the \emph{hard-decision} errors can be alleviated. 
  We first demonstrate the effectiveness of our self-learning methods through ASR experiments in an accent adaptation task between the US and British English speech. Our experiment results show that our method can reduce the WER on the British speech data from 14.55\% to 10.36\% compared to the baseline model trained with the US English data only. Moreover, we investigate the effect of our proposed methods in a federated learning scenario. 

\end{abstract}
\noindent\textbf{Index Terms}: unsupervised learning, self-learning, encoder-decoder, multi-task learning, end-to-end acoustic modeling

\section{Introduction}
\label{sec:intro}
Unsupervised learning of an acoustic model (AM) has a long history from the Bayesian model to the deep neural network (DNN) system in the field of automatic speech recognition (ASR)~\cite{WoodlandPG96,Zavaliagkos98,Lamel02lightlysupervised,Wessel2005,huang2016}. It is typically done by transferring knowledge from stronger teacher model(s) to the student model~\cite{li2014learning,Wong2016,LiSWZG17,Munim2018,Hari2019,MosnerWRPKSMH19,Meng2019} or adapting the seed model pre-trained with a sufficient amount of labeled data~\cite{WoodlandPG96,Zavaliagkos98,Lamel02lightlysupervised,IBMRT2007,KumataniAYMRST12,ManoharGPK18}. Because of no need of heavy decoding processes with teacher models, the latter self-training approach is perhaps preferable in many situations such as on-device personalization and federated learning scenarios~\cite{Konecny15,NIPS2018_7871,Dimitriadis2020,dimitriadis2020federated}. In this work, we thus address the self-training task for AM without an additional bigger teacher model and parallel adaptation data\footnote{The definitions of unsupervised learning and self-learning vary in different applications and fields. For presentation purposes, we will refer any model optimization methods without manual transcripts as unsupervised learning even if the initial seed model is pre-trained with labeled data. We will say self-learning is a special class of unsupervised learning techniques which optimizes the seed model without another teacher model.}. 

The optimization criteria of unsupervised learning for DNN-based ASR systems can fall into two categories: frame-level~\cite{LiSWZG17,Hari2019,MosnerWRPKSMH19,Meng2019}, and sequence-level criteria~\cite{Wong2016,ManoharGPK18,Munim2018}. It is well known that direct sequence-level optimization~\cite{ManoharGPK18} and sequence training from the model initialized with the frame-level cross-entropy objective~\cite{Hari2019,MosnerWRPKSMH19} provide better recognition accuracy. Many successful results have been reported by adapting the hybrid model of the DNN and hidden Markov model (HMM) in an unsupervised way.  

In the hybrid DNN-HMM systems, acoustic, pronunciation, and language models have been trained separately, each with a different objective. While such an approach can use each type of training data efficiently, this disjoint modeling method leads to the sub-optimal solution, especially in terms of model size. This issue has been addressed by designing the \emph{end-to-end} network, which directly generates a word or character sequence from a sequence of speech features. Two popular approaches in this area are perhaps recurrent NN transducer (RNN-T)~\cite{Graves2012RNNT} and attention-based encoder-decoder (AED) network~\cite{ChorowskiNIPS2015_5847,Bahdanau16,ChanJLV15}. Unlike connectionist temporal classification (CTC)~\cite{Graves06connectionisttemporal}, both methods do not make an unreasonable assumption for ASR that the label outputs are conditionally independent of each other. It is shown in~\cite{Prabhavalkar2017} that the AED has the potential of outperforming the RNN-T arguably because of a more flexible assumption on the alignment between input and output, given a large number of training samples~\cite{BattenbergCCCGL17,Chiu2019ACO}. 

Based on the reasons above, we focus on the development of unsupervised learning methods for the AED network. For that, knowledge distillation (KD)~\cite{Hinton2015DistillingTK,KimR16a,Munim2018} is a promising approach. In particular, sequence-level KD~\cite{KimR16a,Munim2018} is more suitable for the AED network to improve modeling capability for sequential input and output. Kim and Rush indeed showed in~\cite{KimR16a} that the sequence-level KD technique can achieve better machine translation performance with lower complexity. Sequence-level KD is also applied to the ASR tasks: model compression~\cite{Munim2018} and domain adaptation with a slight modification of information injected to the student ~\cite{meng2019domain}. Notice that the unsupervised learning method proposed in~\cite{meng2019domain} requires parallel data for the teacher and student models, a pair of clean and noisy data. Again, the use of the bigger teachers or additional parallel adaptation data is not possible in some application scenarios, such as a federated learning scenario~\cite{Dimitriadis2020}.   

In contrast to prior work on unsupervised learning for the AED~\cite{Munim2018,meng2019domain}, we develop the new self-learning techniques which neither require gigantic teacher models nor the parallel data. In such a scenario, we must use the ASR output of the seed model as the target labels. The use of the inaccurate ASR result as the ground truth may degrade recognition accuracy after adaptation. In order to alleviate such a noisy transcript issue, we use the $N$-best hypotheses from the ASR output for adaptation. More specifically, we optimize the AED with multiple objective functions associated with the $N$-best hypotheses. We implement the optimization algorithm with the multi-label learning (MLL) or multi-task learning (MTL) framework~\cite{Caruana1997,Ruder2017AnOO}. Our techniques proposed here are evaluated through two kinds of ASR tasks: an accent adaptation task and a federated learning situation.

The rest of the paper is organized as follows. Section~\ref{sec:backgrond} briefly reviews the sequence-to-sequence model with the AED and its optimization criterion in supervised and unsupervised manners. Section~\ref{sec:proposed} describes our unsupervised learning algorithms for the attention-based encoder-decoder network. Section~\ref{sec:ex} describes the ASR experiments in two kinds of tasks: accent adaptation and federated transfer learning tasks. In section~\ref{sec:conclusions}, we conclude this work.

\section{Background}
\label{sec:backgrond}
\subsection{Attention-based encoder-decoder (AED)}

Let $\bx=[x_1,...,x_K]$ and $\by = [y_1,...,y_M]$ be the feature input and target word embeddings with $K$ and $M$ being the source and target lengths, respectively. ASR involves finding the most probable target sequence given the feature input: 
\begin{equation}
\hat{\by} = \argmax_{\by \in \mathcal{Y}} \log p(\by|\bx)
\lb{eq:decode}
\vspace{-0.1cm}
\end{equation}
where $\mathcal{Y}$ is the set of all possible sequences. As shown in Figure~\ref{fig:las}, $ p(\by|\bx)$ is modeled with an encoder, a decoder and an attention component. The encoder network reads the feature sequence and the decoder network produces a distribution over the target words, one word embedding at a time given the source~\cite{ChanJLV15,BattenbergCCCGL17}. We employ the attention architecture from~\cite{ChorowskiNIPS2015_5847} by replacing the scoring function with the ReLU unit.
\begin{figure}[t]
    \centering
    \includegraphics[width=\linewidth]{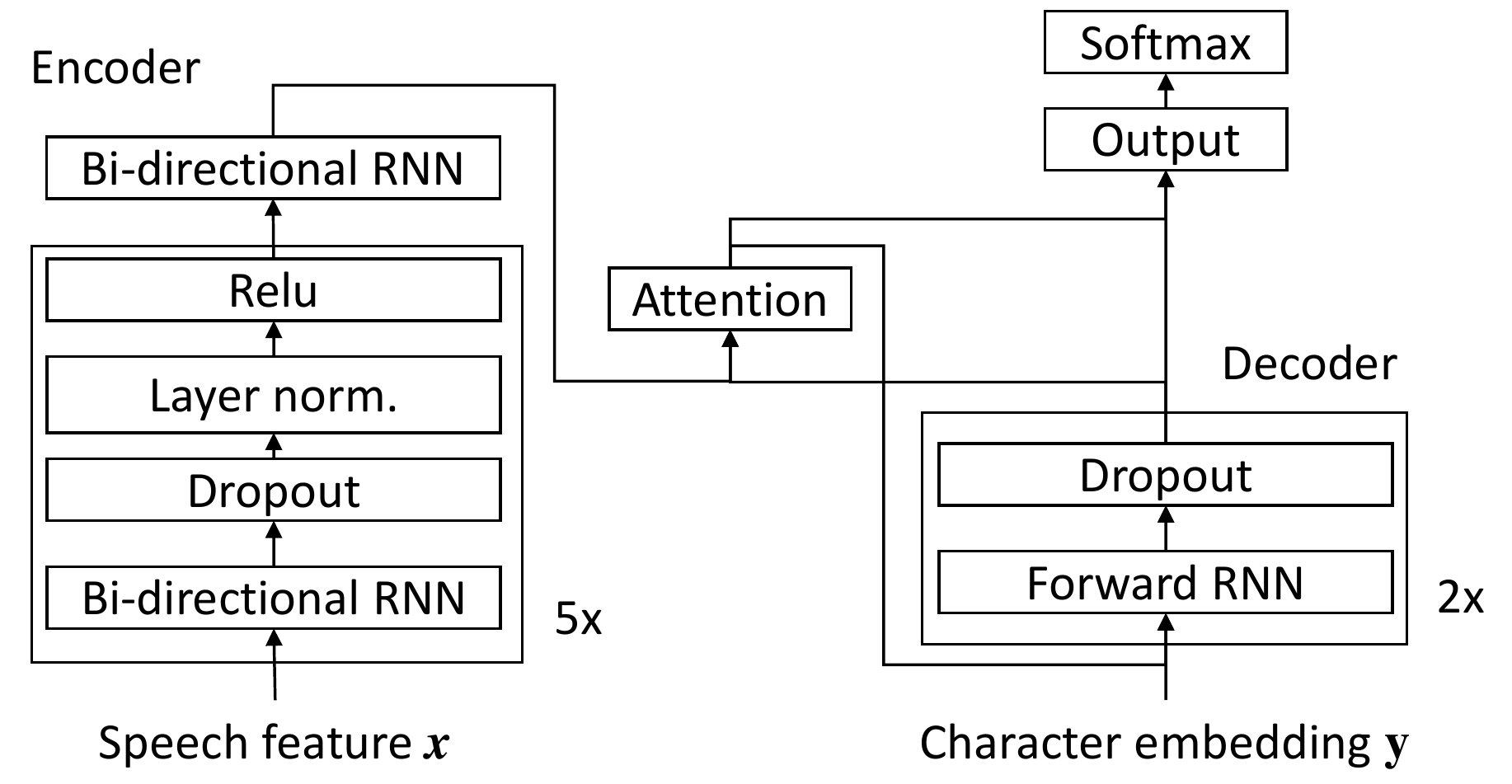}
    \caption{Attention-based encoder-decoder (AED) network}
    \lb{fig:las}
    \vspace{-0.4cm}
\end{figure}

\subsection{Supervised learning}

We briefly review supervised learning for the sequence-to-sequence model before formulating the unsupervised learning problem. First, consider the sequence-level distribution specified with the model parameters $\theta$ over all possible sequences $\by \in \mathcal{Y}$ in the log domain,
\vspace{-0.1cm}
\begin{equation}
\log p(\by | \bx) = \sum_{i=1}^M \log p(y_i | \bx, \by_{<i}; \theta)
\lb{eq:ll}
\vspace{-0.1cm}
\end{equation}
where $\by_{<i}$ is either the ground truth of the previous characters or sampled from the model with a certain sampling rate in order to make the model robust against prediction errors during inference~\cite{ChanJLV15}. The sampling rate can be constant~\cite{ChanJLV15} or scheduled~\cite{BengioVJS15}. For the experiment, we use the scheduled sampling method. Sequence-level training then involves estimating $\theta$ that maximizes the log-likelihood~\erf{eq:ll} over all the sequences. 

\subsection{Unsupervised learning}

The self-learning framework of the attention-based sequence-to-sequence model is very similar with sequence-level knowledge distillation (KD)~\cite{KimR16a}. The only difference is that there is no better teacher model which transfers knowledge to the seed model in the self-learning scenario~\cite{Xie2019SelftrainingWN}. In such a scenario, the seed model must adapt itself to new data without knowledge distillation from the more powerful teacher model; the sequence
distribution of target words can be very noisy. The erroneous labels can degrade accuracy of the seed model severely~\cite{huang2016,Xie2019SelftrainingWN}. In this section, we describe how we approximate the objective function in the case of the unsupervised learning scenario.

First, let us consider the case of the semi-supervised setting with sequence-level KD. As shown in~\cite{KimR16a}, the distribution from the data is replaced with a probability distribution derived from another teacher model $q(\by | \bx)$. The loss function of  sequence-level KD can be written as
\begin{equation}
\mathcal{L}_{\text{SEQ-KD}} = - \sum_{\by \in \mathcal{Y}} q(\by | \bx) \log p (\by | \bx)
\lb{eq:seqkd}
\vspace{-0.1cm}
\end{equation}
Calculating~\erf{eq:seqkd} is computationally intractable. Kim and Rush showed in~\cite{KimR16a} that the loss~\erf{eq:seqkd} could be well approximated with the best hypothesis from the beam search output with the teacher model in a machine translation task. With the first best hypothesis $\hat{\by}_\text{T,1}$, they approximated~\erf{eq:seqkd} as 
\vspace{-0.1cm}
\begin{equation}
\mathcal{L}_{\text{SEQ-KD}} \approx - \log p (\by=\hat{\by}_\text{T,1} | \bx)
\lb{eq:seqkd-approx}
\vspace{-0.1cm}
\end{equation}
It is worth noting that the same algorithm was also applied to an ASR model compression task in~\cite{Munim2018} although their improvement was not as significant as that reported in~\cite{KimR16a}. 

Now consider the case of self-learning. Again, we only have the seed model in this case. Thus, we use the best hypothesis of the seed model $\hat{\by}_\text{S,1}$ instead of that of the teacher model $\hat{\by}_\text{T,1}$. It is straightforward to express the loss approximation function:
\vspace{-0.15cm}
\begin{equation}
\mathcal{L}_{\text{SEQ-KD}} \approx - \log p (\by=\hat{\by}_\text{S,1} | \bx)
\lb{eq:seqkd-approx2}
\vspace{-0.1cm}
\end{equation}
The model can still be improved with self-learning since the seed model learns features from unseen data. However, the seed model needs to provide reasonably good accuracy on the unseen data. Otherwise, the model may degrade due to the erroneous label estimates.

\section{Proposed self-learning method}
\label{sec:proposed} 
\subsection{Loss function approximation with multiple hypotheses}

\begin{figure*}[t]
  \begin{minipage}[t]{.33\linewidth}
    \centering
    \includegraphics[width=\linewidth]{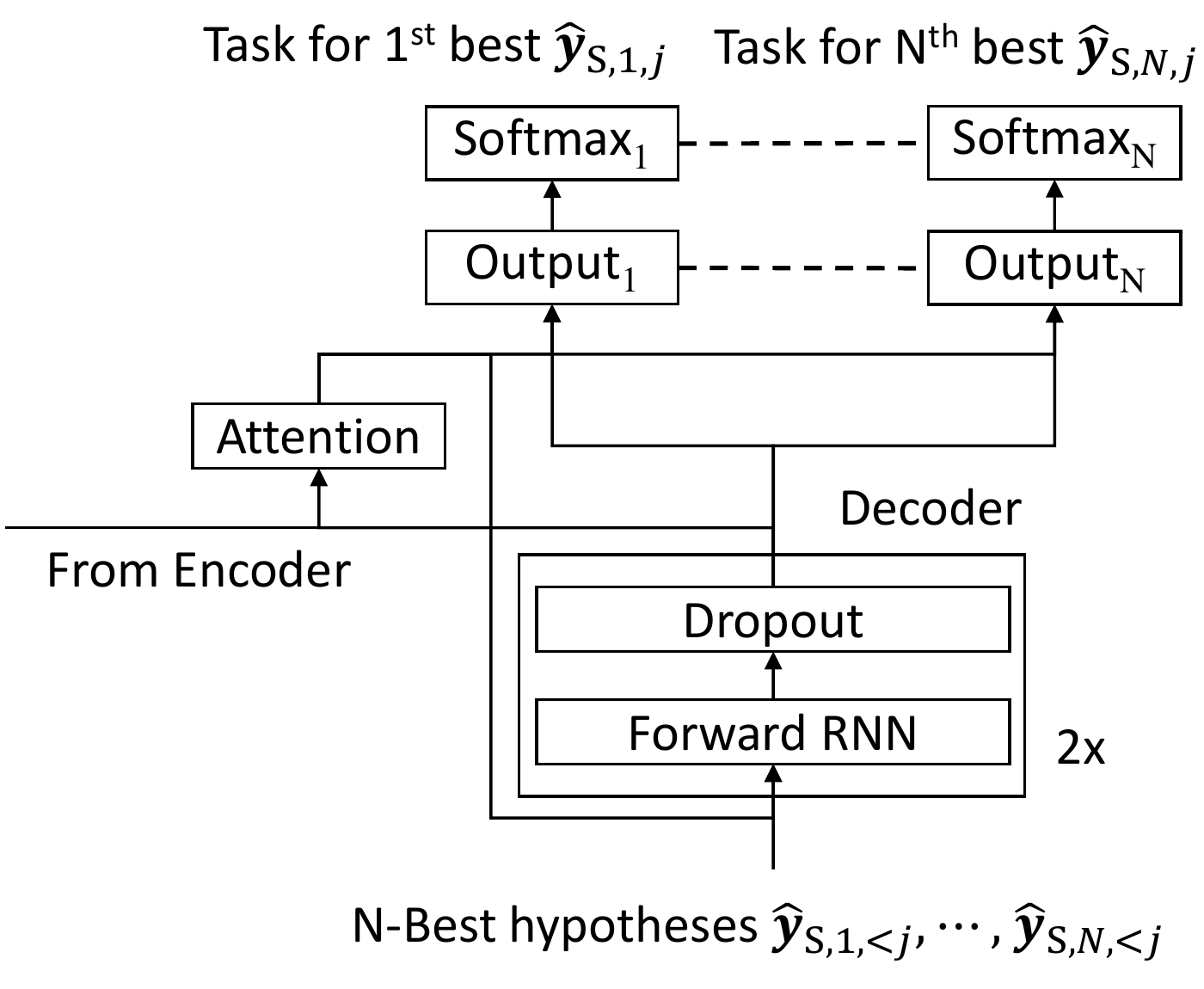}
    \caption{MT network with encoder-decoder shared.}
    \lb{fig:mt-las-wo-enc}
  \end{minipage}\hfill
  \begin{minipage}[t]{.33\linewidth}
    \centering
    \includegraphics[width=\linewidth]{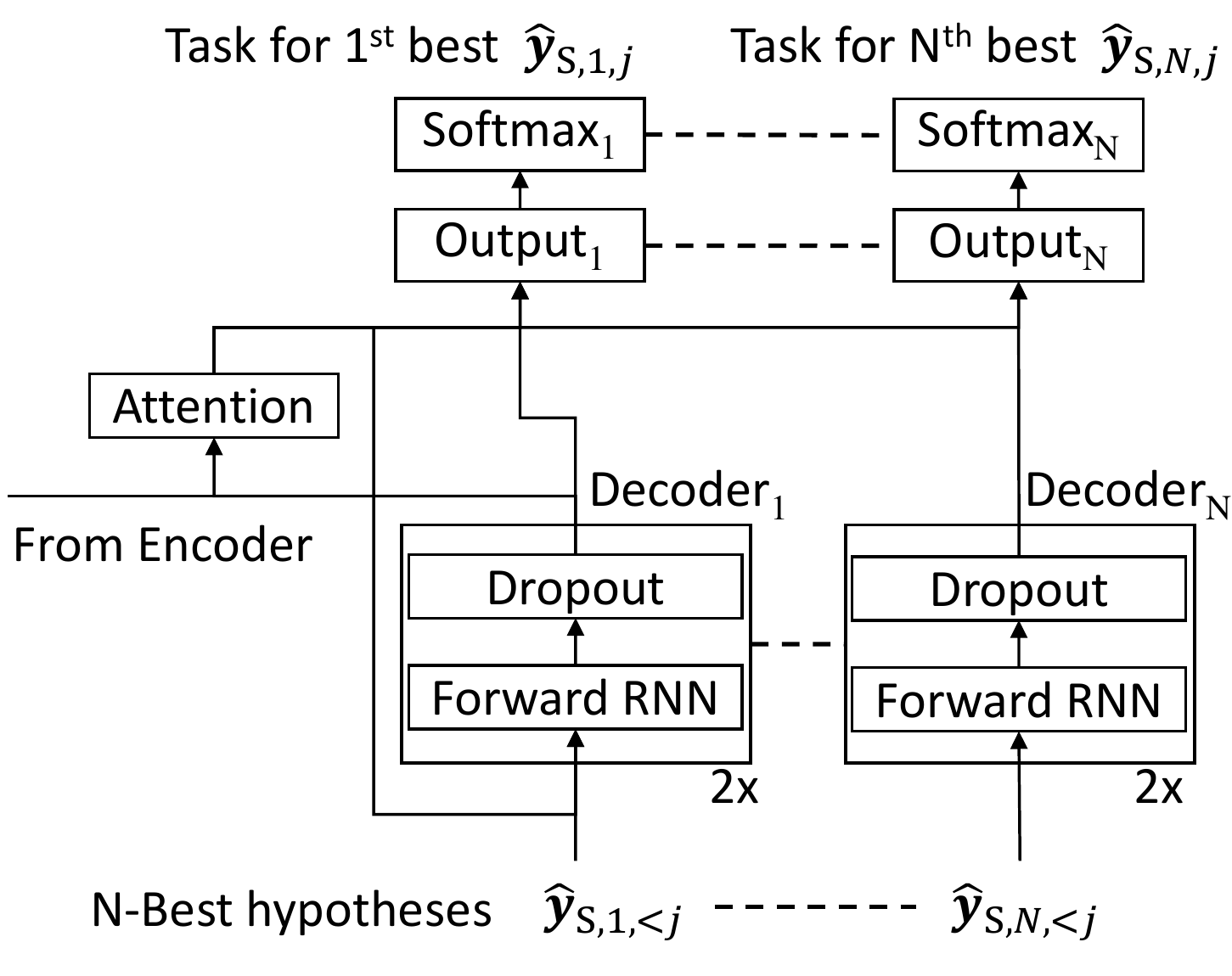}
    \caption{MT network with encoder shared.}
    \lb{fig:mdt-las-wo-enc}
  \end{minipage}\hfill
  \begin{minipage}[t]{.3\linewidth}
    \centering
    \includegraphics[width=\linewidth]{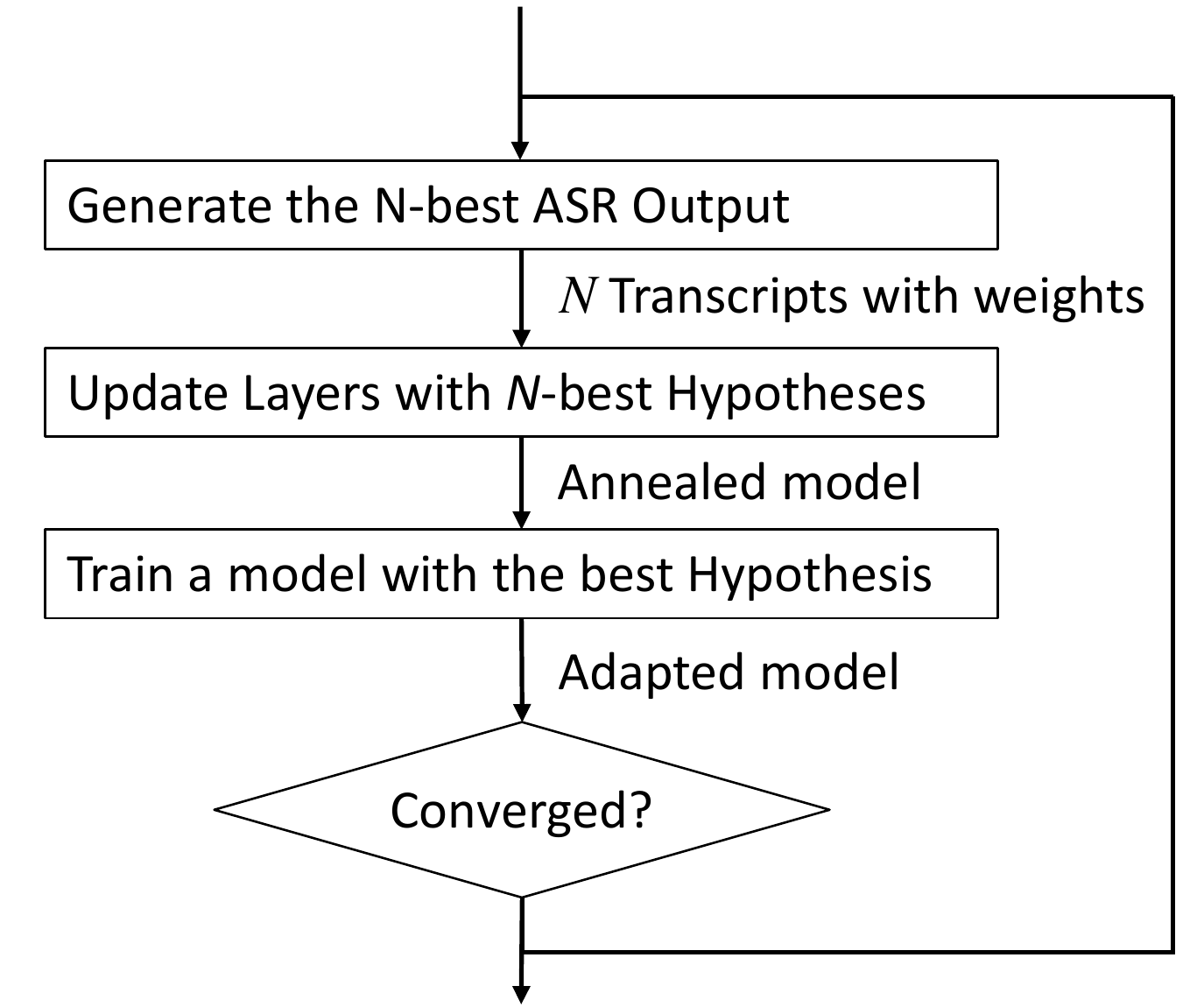}
    \caption{Flow chart of unsupervised training with the $N$-best hypotheses}
    \lb{fig:flow_chart}
  \end{minipage}
  \vspace{-0.4cm}
\end{figure*}

In many cases, the best ASR output contains errors. Thus, approximating~\erf{eq:seqkd} with the only best hypothesis may degrade recognition performance after unsupervised training. Instead, we use the $N$-best hypotheses weighted based on the ASR confidence score for approximation. First, let us denote the weight normalized with the Softmax function for the $n$-th best candidate $\by_{S,n}$ as
\vspace{-0.15cm}
\begin{equation}
q_n(\by_\text{S,n}) = \frac{\exp(s_n/T)}{\sum_{i=1}^{N} \exp(s_i/T)}
\label{eq:asr_weight}
\end{equation}
where $s_n$ is the $n$-th best confidence score and $T$ is a temperature which controls a probability distribution over the classes~\cite{Hinton2015DistillingTK}. We set $T=1$ for the experiments. 
We then rewrite the loss function~\erf{eq:seqkd} as
\begin{equation}
\mathcal{L}_{\text{SEQ-KD}} \approx - \sum_{n=1}^{N} q_n(\hat{\by}_\text{S,n})  \log p (\by=\hat{\by}_\text{S,n} | \bx)
\lb{eq:seqkd-approx3}
\vspace{-0.1cm}
\end{equation}
Minimizing this loss function can be done by creating multiple labels associated with weights for one utterance. We refer to this method as multi-label learning (MLL) in this paper. The loss function can be also optimized through multi-task learning (MTL)~\cite{Caruana1997,Ruder2017AnOO} as it will be explained in the next section. 

\subsection{Optimization with multi-task learning}

Optimization of the loss function~\erf{eq:seqkd-approx3} can be done with the multi-task (MT) network where each separate Softmax layer is associated with the loss of the $n$-th hypothesis. We consider two network architectures in this work. 
Figure~\ref{fig:mt-las-wo-enc} shows the first architecture proposed here. In the architecture shown in Figure~\ref{fig:mt-las-wo-enc}, the parameters of the encoder, attention and decoder network are shared during the optimization process. Only the output layer is separated for each task associated with the term in~\erf{eq:seqkd-approx3}. The weights of the new branch for the $n$-th best hypothesis are initialized with those of the main task. 

Figure~\ref{fig:mdt-las-wo-enc} shows the second architecture investigated in this work. In contrast to the architecture in Figure~\ref{fig:mt-las-wo-enc}, the second network architecture shares the encoder and attention module only; the set of the decoder and the output layer is separated for each task. After self-training, the encoder is expected to generate a better acoustic embedding that covers the multiple hypotheses. This can be viewed as acoustic feature-space adaptation with the AED network. 

Figure~\ref{fig:flow_chart} shows a flow chart of our unsupervised training scheme with the $N$-best hypotheses. As shown in Figure~\ref{fig:flow_chart}, we first decode the sequence of the speech features with the beam search in order to obtain the $N$-best hypotheses, $\hat{\by}_\text{S,1}$~$\cdots$~$\hat{\by}_\text{S,N}$. Then, the seed model is adapted with the $N$ hypotheses. This can be done by either multi-label learning (MLL) or MTL framework, as described before. The normalized probability computed with~\erf{eq:asr_weight} is used for weighting the loss of each utterance in MLL and the task in MTL. After adopting the model with the $N$-th hypotheses, we train the model with the best hypothesis only. The only single task branch is used in the case of MTL. These steps will be repeated until there is no improvement in a character error rate on the validation data. During decoding, we can remove the sub-task branches and keep the main task branch only.

\section{ASR experiment}
\label{sec:ex} 
In this section, we describe the ASR experiments in order to compare the unsupervised learning techniques with the AED. 

\subsection{Accent Adaptation Task}
\label{sec:ex1}

First, we addressed the English accent adaptation task from American English to British English. The data used for the experiment were collected through the live traffic and stored with the sampling rate of 16~kHz. For the experiments, we train the seed model with approximately 75,000 hours of American English speech data. About 1,000 hours of the British English speech data were used for adaptation. The adaptation data are treated as untranscribed in the case of unsupervised learning experiments. The British Cortana corpus was used as a test set. The test data were also collected through the live traffic and consists of 14,000 utterances spoken by hundreds of users. 

The feature extraction front-end computes 40 log mel-filterbank energy features at a frame rate of 10 milliseconds (msec), concatenates three adjacent feature frames, and then downsamples it into a 30 msec frame rate. Each feature was rescaled to have zero mean and unit variance over the training set. The global mean and variance calculated with the training data were applied to adaptation and evaluation. In order to augment adaptation data, we randomly scale the amplitude of a signal, change the speed in the same manner implemented in Sox software~\cite{Sox} and perform the spec-augment on the feature domain~\cite{Park2019}. The network used here has 6 layers of 1024 bidirectional GRU nodes in the encoder and 2 layers of 1024 unidirectional GRU nodes in the decoder. All dropout types were applied with $0.1$ dropout probability. We used label smoothing and Adam optimizer in the experiments.

Table~\ref{tab:asr-results1} shows the word error rates (WERs) obtained with different models, the seed model trained with American English data, British AMs adapted in the supervised and unsupervised learning manners. In the self-learning scenario, we performed 4 types of sequence-level self-knowledge distillation methods, conventional training with the 1-best hypothesis~\erf{eq:seqkd-approx2}, multi-label learning (MLL) with the 4-best hypotheses~\erf{eq:seqkd-approx3}, multi-task learning (MTL) with the attention, encoder and decoder (AED) shared and that with the attention and encoder (AE) shared. For unsupervised MTL, we use the 4-best hypotheses. It is clear from Table~\ref{tab:asr-results1} that the WER of the American English model can be reduced from 14.55\% to 7.8\% by supervised learning with the manual transcripts. It is also clear from Table~\ref{tab:asr-results1} that the WER of the seed model can be reduced by the self-learning methods, although the improvement is not as significant as that obtained with supervised learning. Table~\ref{tab:asr-results1} also shows that the use of the 4-best hypotheses provides better recognition accuracy than the case of the 1-best hypothesis only. The WERs of the 1-best and N-best results on the adaptation data obtained with the seed model were 22.6\% and 13.7\%, respectively. Thus, we consider using multiple hypotheses to provide better recognition accuracy. We did not observe a significant difference among different self-learning methods with the 4-best hypotheses after decoding the adaptation data twice, but it is worth mentioning that MTL with the shared AE network provided the lowest WER with the first decoding pass. 

\begin{table}
\begin{center}
{\scriptsize
\begin{tabular}{|c|c||c|}
\hline
Adaptation Type & Adaptation Method & WER(\%) \\
\hline
\hline
Seed model & US baseline model   & 14.55 \\
\hline
Supervised learning & Training with transcription & 7.8 \\
\hline
 \multirow{4}{*}{Self-learning}& Training with the 1-best hypo & 12.09 \\
 & MLL with 4-best hypos & 10.66 \\
    & MTL with shared AED & 10.83 \\
    & MTL with shared AE  & 10.36 \\
\hline
\end{tabular}
}
\end{center}
\vspace{-0.1cm}
\caption{Word error rates (WER) on the British Cortana data for each training method}
\lb{tab:asr-results1}
\vspace{-0.8cm}
\end{table}

\subsection{Librispeech Task}
\label{sec:ex2}
We also conducted ASR experiments with the Librispeech (LS) data~\cite{Panayotov2015LibrispeechAA} for this study. For the LS task, we used the same front-end as described in Section~\ref{sec:ex1} but slightly different model configuration. The AED network used here consists of 6 layers of 1024 bidirectional LSTM nodes and 2 layers of 1024 unidirectional LSTM nodes. For the LS task, we used byte-pair encoding (BPE)~\cite{Sennrich2016BPE} to create 16,000 subword units. The optimizer settings are the same as described in~\cite{Dimitriadis2020}. 

The LS corpus contains approximately 1000 hours of read speech for training. We split the whole LS training set into two subsets. A seed model was first trained with the first half of the training set. For adaptation, we further divide the second half dataset into the per-speaker dataset. The adaptation dataset contains approximately 1100 speakers. Since the seed model does not see the LS vocabulary and word sequences of the adaptation dataset, the ASR result on the adaptation data will not be perfect; self-learning has to be done with noisy labels. The seed model was then adapted with the speaker-partitioned dataset through our federated learning simulator~\cite{Dimitriadis2020,dimitriadis2020federated}. Briefly, the federated learning algorithm iteratively performs the following processes on a server and clients: (1) broadcasting the global (seed) model from the server to the clients, (2) decoding speech with the global model at each client, (3) adapting the global model with each client's data and the ASR hypotheses, (4) encrypting each client's model so as to ensure privacy\footnote{We did not use encryption here for the sake of simplicity although it is straightforward to apply multi-party computation (MPC) secure algorithms or differential privacy techniques such as~\cite{NIPS2018_7871}.}, (5) aggregating the linearly-weighted models from a pool of clients, and (6) optimizing the global model through the aggregated gradients. Those processes are repeated until the loss on the validation data is converged. We only decode speech every 512 aggregation steps for unsupervised training. We found that this could avoid training divergence due to inaccurate ASR results with an intermediate model trained halfway. 
 Notice that the adapted model is only uploaded to the server while privacy data such as speech and transcripts are secured in the client. The adapted models will be sent to the server after processing all the utterances per speaker on each client. This saves a tremendous amount of network bandwidths. 

Table~\ref{tab:asr-results2} shows the WERs on the LS test clean set obtained with different models in the same way as Table~\ref {tab:asr-results1}. We performed supervised and unsupervised learning algorithms on the seed model with the second half of the training data. The last row in Table~\ref{tab:asr-results2} shows the upper bound WER in the case that the model is trained from scratch with the full amount of the LS training data.  It is clear from Table~\ref{tab:asr-results2} that the WER of the seed model can be significantly reduced by supervised-training with the rest 50\% of the training data from 5.66\%to 4.40\%. It is also clear from Table~\ref{tab:asr-results2} that recognition accuracy can still be improved by self-learning with MTL even when the seed model does not know the vocabulary and word sequences of the adaptation data. We observed that the validation loss diverged in MLL, which very likely led to recognition accuracy degradation. The results also show that the convergence performance is improved by separating more layers relevant to the language model, the decoder and output layers, for each hypothesis. This suggests that the $N$ best hypotheses contain little language information shareable over the decoders and output layers. This may be because language knowledge of the seed model on the adaptation dataset is poor. 
In order to check how much the N-best results can potentially help recognition accuracy, we decoded the second half of the LS training data with our LS seed model.  The WERs of the 1-best and N-best results on the LS adaptation data were 10.4\% and 7.62\%. The difference between the 1-best and N-best results was not as significant as that in the case of the accent adaptation task. We assume that many word errors in this LS task cannot be recovered by simply increasing the $N$-best hypotheses because the seed model does not observe the language of the adaptation dataset. On the other hand, the seed model used in the British Cortana task should cover the vocabulary and grammar for the adaptation data very well because it is trained with a large amount of the American English data containing the Cortana task. Of course, there are some different word usages between the American and British English. We are, however, led to conclude that the short phrase ASR task such as the Cortana task was probably not affected significantly. 

\begin{table}
\begin{center}
{\scriptsize
\begin{tabular}{|c|c||c|}
\hline
Adaptation Type & Adaptation Method & WER(\%) \\
\hline
\hline
Seed model & Trained from scratch with 50\% of data  & 5.66 \\
\hline
Supervised learning & Training with the rest 50\% of data & 4.40 \\
\hline
 \multirow{4}{*}{Self-learning}& Training with the 1-best hypo &  5.58 \\
 & MLL with 4-best hypos &  5.72 \\
    & MTL with shared AED & 5.55 \\
    & MTL with shared AE  & 5.22 \\
\hline
\multirow{2}{*}{Reference}& The whole LS training data & 4.00 \\
& in centralized training & \\
\hline
\end{tabular}
}
\end{center}
\vspace{-0.1cm}
\caption{WERs on the LS test clean dataset for each training method in the privacy preserving scenario (where the seed model has no language information on the adaptation data)}
\lb{tab:asr-results2}
\vspace{-0.9cm}
\end{table}

%
%

\section{Conclusions}
\label{sec:conclusions}

In this work, we have investigated the ASR accuracy of self-learning methods with the attention-based sequence-to-sequence model. We proposed the new self-learning methods with the $N$ best hypotheses; the seed model is adapted with MLL or MTL in an end-to-end manner. We conducted two kinds of ASR experiments: the English accent adaptation task and the federated-learning scenario. In the accent adaptation task, our experiment results show that our best self-learning method can reduce the WER on the British speech data from 14.55\% to 10.36\% compared to the baseline model trained with the US English data only. In the federated-learning scenario, the improvement on the LS dataset was not as significant as that on the Cortana dataset. In contrast to the Cortana task, the seed model used in the LS task does not contain language information on the adaptation data well. Thus, misrecognition results may not be entirely recoverable by increasing the N-best candidates. We plan to combine our AM self-learning and the language data augmentation method a text-to-speech module~\cite{Ney2020}

\section{Acknowledgements}
The authors would like to thank Masaki Itagaki, Heiko Rahmel, Naoyuki Kanda, Lei He, Ziad Al Bawab, Jian Wu, and Xuedong Huang for their project support and technical discussions.

\bibliographystyle{IEEEtran}

\bibliography{mybib}
\end{document}